\documentclass[11pt,a4paper]{article}

\usepackage[margin=22mm]{geometry}
\usepackage{graphicx}
\usepackage{adjustbox}
\usepackage{booktabs}
\usepackage{amsmath,amssymb}
\usepackage{hyperref}
\usepackage{caption}
\usepackage{subcaption}
\usepackage{microtype}
\usepackage{url}
\usepackage[table]{xcolor}
\definecolor{tablebest}{RGB}{212,237,218}
\usepackage{placeins}
\usepackage{float}
\usepackage{tikz}
\usetikzlibrary{shapes.geometric,arrows.meta,positioning,calc,fit,backgrounds}

\graphicspath{{figures/}}

\hypersetup{colorlinks=true,linkcolor=blue,citecolor=blue,urlcolor=blue}

\title{%
  \textbf{Device-First Feedback: Toward Mobile-Native}\\
  \textbf{LLM-Driven Neural Architecture Search}%
}

\author{%
  Saif~U~Din
  \and Muhammad~Ahsan~Hussain
  \and Radu~Timofte
  \and Dmitry~Ignatov
  \and \\[0.66em]
  Computer Vision Lab, CAIDAS \& IFI, University of W\"urzburg, Germany%
}

\date{}

\begin{document}
\maketitle

\begin{abstract}
Deploying large-language-model (LLM) generated convolutional neural networks on
real mobile hardware requires more than GPU validation accuracy: INT8 TensorFlow
Lite export, delegate selection, and on-device latency jointly determine
whether a model is usable.
We present an \emph{automated mobile deployment pipeline} that closes the loop
from QLoRA fine-tuning of an architecture-generating LLM~\cite{kochnev2025nngpt}
through GPU evaluation, INT8 export, and physical-device benchmarking to gated
augmentation of the training corpus.
The pipeline is fully scripted (\texttt{TuneNNGen --mobile\_deployment}) and
runs cycle-by-cycle without manual intervention, with resume support after
interruptions.
We evaluate the \textbf{same frozen protocol} on two benchmarks---CIFAR-10 and
CIFAR-100~\cite{krizhevsky2009learning} on a Samsung SM-P613 tablet (seed~42,
20 models per cycle, cycles~0--6).
On CIFAR-10, cycle~1 gate-accepts and improves the mobile deployment score
$\approx 25.6\times$ over baseline with mean quantized accuracy 46.9\%;
later cycles raise GPU accuracy but fail the non-decreasing mobile gate.
On CIFAR-100, the pre-QLoRA baseline retains the best mobile score; iterative
rounds improve GPU accuracy (up to 26.2\%) yet cannot surpass cycle~0 on-device,
and the training pool stalls at 19 examples after the first accepted round.
Together, the two studies show that closed-loop GPU fine-tuning does not
guarantee monotonic mobile gains---especially on harder classification---and
that multi-dataset, on-device measurement is needed to stress-test deployment
objectives.
We release per-cycle metrics with 95\% confidence intervals, all figures, and
complete reproduction commands.
\end{abstract}

\section{Introduction}
\label{sec:intro}

Neural architecture search and LLM-based code generation can propose diverse
CNN topologies quickly~\cite{kochnev2025nngpt,brown2020language}, but edge deployment
adds constraints that workstation GPU accuracy alone does not capture:
post-training INT8 quantization shifts accuracy~\cite{jacob2018quantization}, and
NNAPI/GPU/CPU delegates exhibit heterogeneous latency~\cite{zhang2021nnmeter,reddi2020mlperf}.

Prior mobile benchmarks~\cite{tan2019mnasnet,reddi2020mlperf} measure fixed model
zoos offline; they do not address \emph{iterative} improvement of a generative
model using real device feedback.
We automate a \textbf{closed-loop mobile deployment pipeline} that, each cycle:
(1)~fine-tunes an LLM on architecture chat data (QLoRA~\cite{dettmers2023qlora});
(2)~generates $K$ candidate \texttt{Net} modules in LEMUR-compatible format;
(3)~trains each candidate for one GPU epoch on the target dataset (CIFAR-10 or CIFAR-100);
(4)~exports INT8 TFLite and benchmarks on a physical Android device;
(5)~selects novel models by mobile score and augments training data only if a
\textbf{non-decreasing gate} accepts the cycle.

\paragraph{Contributions.}
\begin{itemize}
  \item An end-to-end \textbf{automated pipeline} (Figure~\ref{fig:workflow-complete},
        \S\ref{sec:pipeline}) from LLM fine-tuning through on-device measurement, invocable as a single
        command and resumable per cycle (\S\ref{sec:running}).
  \item A \textbf{mobile-first selection metric}
        $\textit{score} = \textit{quantized\_accuracy} / \textit{latency}$ with
        delegate-aware benchmarking on SM-P613.
  \item \textbf{Two reproducible benchmarks} (CIFAR-10 and CIFAR-100, cycles~0--6,
        seed~42) with per-cycle 95\% confidence intervals, cross-dataset comparison
        (Table~\ref{tab:cross-dataset}), and publication-ready figures
        (\S\ref{sec:results-cifar10}--\S\ref{sec:results-cifar100}).
  \item Analysis of \textbf{GPU--mobile divergence}: GPU accuracy can rise while
        mobile score falls; outcomes differ by dataset difficulty, gate history, and
        training-pool growth (190 vs.\ 19 examples).
\end{itemize}

\section{Related Work}
\label{sec:related}

\paragraph{Hardware-aware architecture design.}
Designing convolutional networks for mobile hardware has motivated depthwise
separable architectures~\cite{howard2017mobilenets} and, subsequently, neural
architecture search with latency as a first-class objective.
MnasNet~\cite{tan2019mnasnet} optimizes expected latency on representative
platforms through a search controller and surrogate costs, while
ProxylessNAS~\cite{cai2018proxylessnas} searches directly on the target task
with on-device latency estimates.
These methods operate over a fixed search space or supernet and emit a single
architecture per deployment target.
In contrast, our generator is a language model that emits full training
specifications, and hardware feedback determines which generations enter its
fine-tuning corpus; the object of optimization is the \emph{generator}, not an
individual architecture.

\paragraph{On-device inference measurement.}
Reproducible edge benchmarking requires agreed metrics and runtimes.
MLPerf Inference~\cite{reddi2020mlperf} standardizes latency and throughput
reporting across devices, and nn-Meter~\cite{zhang2021nnmeter} predicts latency
from computation graphs when exhaustive profiling is impractical.
On Android, TensorFlow Lite~\cite{tensorflowlite} exposes multiple execution
delegates (CPU, GPU, NNAPI), so the same quantized graph can exhibit different
accuracy--latency trade-offs depending on the backend.
Existing studies score \emph{static} model zoos offline; our pipeline instead
treats on-device measurements as training-time signals, ranking candidates by a
deployment objective and gating corpus growth on it.

\paragraph{Quantization and the deployment gap.}
Post-training and quantization-aware training enable integer-only
inference~\cite{jacob2018quantization}, but quantization interacts nonlinearly
with architectural choices: models that train well in floating point may lose
accuracy or run slowly after INT8 conversion and delegate lowering, so
workstation validation accuracy is an incomplete proxy for edge usability.
Our protocol makes this gap explicit by exporting every candidate to INT8,
measuring quantized accuracy and physical-device latency each cycle, and
accepting generative-model updates only when the deployment score does not
regress; the observed divergence between GPU accuracy and mobile score
(\S\ref{sec:results-compare}) empirically confirms the concern.

\paragraph{LLM-driven architecture generation.}
Large language models generate code from natural-language and few-shot
prompts~\cite{brown2020language}, and parameter-efficient adaptation
(LoRA~\cite{hu2022lora}, QLoRA~\cite{dettmers2023qlora}) makes repeated
fine-tuning of billion-parameter models feasible on academic hardware.
NNGPT~\cite{kochnev2025nngpt} frames the LLM as a self-improving AutoML engine
that generates, trains, and learns from executable PyTorch pipelines on the
LEMUR corpus; retrieval-augmented variants such as
\texttt{ABrain/NNGPT-UniqueArch-Rag}~\cite{nngpt_uniquearch_rag} ground
synthesis in verified code.
Prior work in this family evaluates in the source (PyTorch) domain.
We extend the paradigm to the post-export regime: device-resolved INT8 metrics
enter the same iterative loop as GPU evaluation, with an explicit acceptance
rule on the deployment score.
To our knowledge, the combination of fully scripted multi-cycle QLoRA, physical
Android benchmarking, and gated data augmentation---evaluated on paired
standard benchmarks~\cite{krizhevsky2009learning}---has not previously been
reported as an end-to-end system.

\FloatBarrier
\section{Automated Mobile Deployment Pipeline}
\label{sec:pipeline}

\begin{figure}[tbp]
\centering
\adjustbox{max width=\textwidth,center}{%
\begin{tikzpicture}[
  font=\scriptsize,
  node distance=4mm and 5mm,
  box/.style={rectangle, rounded corners=2pt, draw=black!60, align=center,
              minimum height=7mm, inner sep=2pt, text width=14mm, fill=blue!6},
  mob/.style={rectangle, rounded corners=2pt, draw=black!60, align=center,
              minimum height=7mm, inner sep=2pt, text width=14mm, fill=orange!14},
  init/.style={rectangle, rounded corners=2pt, draw=black!60, align=center,
               minimum height=7mm, inner sep=2pt, text width=13mm, fill=purple!8},
  gate/.style={diamond, aspect=2.2, draw=black!60, align=center, inner sep=0.5pt,
               fill=yellow!18, font=\tiny},
  data/.style={cylinder, shape border rotate=90, aspect=0.22, draw=black!60,
               align=center, minimum height=8.5mm, minimum width=12mm, fill=green!10},
  arr/.style={-{Latex[length=1.6mm]}, semithick, draw=black!70},
  darr/.style={-{Latex[length=1.6mm]}, semithick, draw=black!70, dashed}
]

\node[init] (lemur) {LEMUR\\seeds};
\node[init, right=of lemur] (allow) {INT8\\allowlist};
\node[data, right=of allow] (chat0) {Mobile\\seed pool};
\node[box, right=of chat0] (val) {Validate\\prereqs};
\draw[arr] (lemur)--(allow)--(chat0)--(val);
\node[font=\tiny\itshape, above=1mm of allow] {One-time bootstrap};

\node[box, below=8mm of lemur] (base) {Base LLM\\(no adapter)};
\node[box, right=of base] (g0) {Generate\\$K$ nets};
\node[box, right=of g0] (e0) {GPU\\1 epoch};
\node[mob, right=of e0] (m0) {INT8 +\\device};
\node[data, right=of m0] (ref) {Cycle~0\\ref.};
\draw[arr] (base)--(g0)--(e0)--(m0)--(ref);
\draw[darr] (val.south) -- ++(0,-2mm) -| (base.north);
\node[font=\tiny\itshape, left=0mm of base] {\texttt{--baseline\_only}};

\node[data, below=9mm of base] (dc) {Train pool\\$\mathcal{D}_{c-1}$};
\node[box, right=of dc] (ft) {QLoRA\\3 epochs};
\node[box, right=of ft] (gen) {Generate\\$K$ nets};
\node[box, right=of gen] (gpu) {GPU train\\+ filter};
\node[mob, right=of gpu] (tfl) {INT8\\TFLite};
\node[mob, right=of tfl] (dev) {SM-P613\\bench};

\coordinate (bus) at ([xshift=-11mm]dc.west);

\draw[arr] (dc)--(ft)--(gen)--(gpu)--(tfl)--(dev);

\node[data, below=7mm of dc] (poolc) {$\mathcal{D}_c$\\train.jsonl};
\node[box, right=of poolc] (aug) {Append to\\train pool};
\node[box, right=of aug] (tochat) {Convert\\to chat};
\node[gate, right=of tochat] (gated) {Gate\\ok?};
\node[mob, right=of gated] (sel) {Score\\+ novelty};

\draw[arr] (dev.south) -- (dev.south |- sel.north) -- (sel.north);
\draw[arr] (sel)--(gated);
\draw[arr] (gated)-- node[above, font=\tiny] {yes} (tochat)--(aug);
\draw[arr] (aug.west) -- (poolc.east);
\draw[darr] (gated.west) -- ++(-3mm,0) coordinate (gw)
  (gw |- poolc.east) -- (poolc.east)
  node[pos=0.35, above, font=\tiny] {no};

\draw[arr] (poolc.north) -- node[left, font=\tiny, align=right] {next\\cycle} (dc.south);

\draw[darr] (chat0.west) -- (bus |- chat0.west) -- (bus |- dc.north) -- (dc.north);

\draw[darr] (ref.south) -- (ref.south |- gated.east) -- (gated.east);

\begin{scope}[on background layer]
  \node[draw, dotted, rounded corners, inner sep=3pt, fill=purple!3, fit=(lemur)(val)] {};
  \node[draw, dotted, rounded corners, inner sep=3pt, fill=blue!3, fit=(base)(ref)] {};
  \node[draw, dotted, rounded corners, inner sep=3pt, fill=orange!5,
        fit=(dc)(dev)(sel)] {};
  \node[draw, dotted, rounded corners, inner sep=3pt, fill=green!8,
        fit=(gated)(poolc)(tochat)(aug)] {};
\end{scope}

\end{tikzpicture}%
}
\caption{Complete workflow of the automated mobile deployment pipeline
(\texttt{TuneNNGen --mobile\_deployment}).
\textbf{Top:} bootstrap and optional cycle~0 baseline.
\textbf{Middle:} one iterative cycle---train pool $\mathcal{D}_{c-1}$, QLoRA,
generation, GPU evaluation, INT8 export, and on-device benchmark (horizontal
chain only).
\textbf{Bottom:} selection, gate, conversion, and append to
\texttt{train.jsonl}; $\mathcal{D}_c$ sits under the train-pool column and
feeds the next cycle via a vertical link (no path crosses QLoRA).
If the gate rejects, dashed flow skips new rows but the pool file is carried
forward unchanged.}
\label{fig:workflow-complete}
\end{figure}
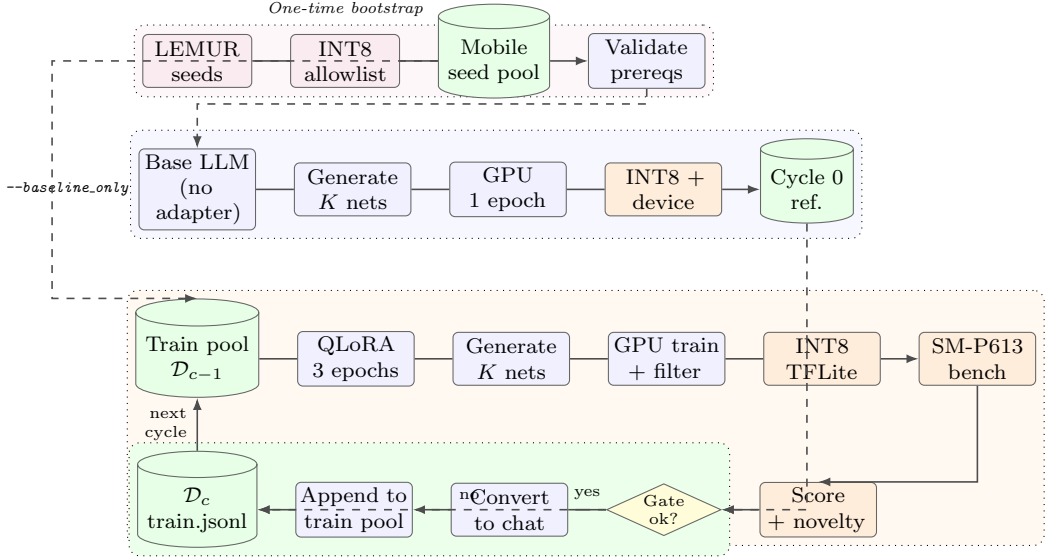

Figure~\ref{fig:workflow-complete} shows the end-to-end workflow: a one-time
bootstrap, an optional pre-fine-tuning baseline (cycle~0), and the gated
iterative loop for cycles $c \geq 1$.
The orchestrator, \texttt{MobileDeploymentFinetuner}, is invoked through
\texttt{TuneNNGen} when \texttt{--mobile\_deployment} is set; every cycle is
persisted to disk (\texttt{cycle\_$c$/cycle\_results.json} together with TFLite
graphs, weights, and Android benchmark records), which makes the pipeline
resumable and every reported number regenerable from raw artifacts.

\subsection{Initialization and baseline}
\label{sec:init}

On first launch the pipeline prepares a mobile-conditioned seed corpus.
If no chat data exists, the standard LEMUR curation step builds architecture
chat examples from the accepted-code corpus; these are then intersected with an
INT8 deployability allowlist so that only architectures with a verified
TensorFlow Lite export path remain (for CIFAR-100 the architecture catalog of
the CIFAR-10 allowlist is reused, while all training and evaluation use 100
classes).
Known quantized accuracies and model names from the allowlist are injected into
the seed prompts, so the language model conditions on realistic mobile
statistics from the start.
A manifest records which rows were kept, and an empty filter result is rejected
rather than allowed to wipe the corpus.
Finally, prerequisites (GPU availability, disk space, LLM configuration, and
non-empty training data) are validated before any cycle starts.

An optional baseline cycle (cycle~0, \texttt{--baseline\_only}) evaluates the
pristine base model \texttt{ABrain/NNGPT-UniqueArch-Rag} without any adapter:
$K{=}20$ architectures are generated from the mobile prompt template, each
syntactically valid model is trained for one GPU epoch, and every successful
model is exported to INT8 TFLite and benchmarked on the physical device.
Baseline models never augment the training pool and the gate is disabled for
this cycle; its purpose is to establish the pre-fine-tuning deployment
reference against which later cycles are compared.

\subsection{Iterative cycle}
\label{sec:cycle}

Each cycle $c \geq 1$ executes the following stages without manual
intervention; stages whose outputs already exist on disk are skipped, which
yields idempotent resume behavior.

\begin{enumerate}
  \item \emph{Training data.} Cycle~1 reads the seed corpus; cycle $c{>}1$
        reads the pool $\mathcal{D}_{c-1}$ produced by the previous
        augmentation step.
  \item \emph{QLoRA fine-tuning.} The LLM is fine-tuned for three epochs on
        $\mathcal{D}_{c-1}$ using QLoRA~\cite{dettmers2023qlora}; the resulting
        adapter is stored in an isolated per-cycle checkpoint.
  \item \emph{Generation.} The cycle adapter generates up to $K{=}20$ new
        \texttt{Net} modules; a global checksum prevents duplicate model
        identifiers across cycles.
  \item \emph{GPU evaluation.} Each accepted architecture is trained for one
        epoch on the active dataset, and models below the dataset-specific
        accuracy threshold are dropped; trained weights are retained for
        faithful mobile export.
  \item \emph{Mobile stage.} Every GPU-successful model is converted to INT8
        TFLite by post-training quantization, its quantized top-1 accuracy
        $a_i^{\mathrm{int8}}$ is measured on a 256-image holdout, and latency
        is benchmarked on the device for the CPU, GPU, and NNAPI delegates
        (20 timed runs each); the minimum median duration
        $t_i^{\mathrm{best}}$ across successful delegates is recorded.
  \item \emph{Selection and gate.} Models with valid mobile metrics are ranked
        by the deployment score of Eq.~\eqref{eq:mobile-score} and passed
        through hash-based novelty filtering; the cycle is accepted only if
        $s_c^{\max} \geq 0.99 \cdot s_{\mathrm{ref}}$ and at least two models
        have valid scores, where $s_c^{\max}$ is the best score in cycle $c$
        and $s_{\mathrm{ref}}$ is the best score among prior gate-eligible
        cycles (including the cycle~0 reference for $c \geq 2$).
  \item \emph{Augmentation.} For accepted cycles, selected models are converted
        to chat examples whose prompts embed both GPU accuracy and mobile
        metrics, and appended to form $\mathcal{D}_{c}$; for rejected cycles no
        new rows are added and the pool is carried forward unchanged.
\end{enumerate}

\subsection{Mobile deployment score}
\label{sec:mobile-score}

Candidates are ranked by a single deployment-centric objective,
\begin{equation}
  \textit{score}_i \;=\; \frac{a_i^{\mathrm{int8}}}{t_i^{\mathrm{best}}},
  \label{eq:mobile-score}
\end{equation}
where $a_i^{\mathrm{int8}}$ is the INT8 TFLite top-1 accuracy on the active
dataset's test split and $t_i^{\mathrm{best}}$ is the minimum median latency in
milliseconds among successful delegate runs; higher is better.
Because the gate enforces a near-non-decreasing best score across cycles,
iterative fine-tuning cannot silently regress the deployment objective even
when workstation metrics improve.

\subsection{Implementation and reproducibility}
\label{sec:running}

The driver \texttt{TuneNNGen} constructs the finetuner, loops over cycles, and
aggregates per-cycle results into a single summary; a companion plotting script
rebuilds all figures and the metrics JSON bundled with this paper from the
stored cycle records, once per curation root (CIFAR-10 and CIFAR-100 runs are
kept in separate artifact trees).
The complete command lines for the baseline cycle, full iterative runs on
either dataset, resuming after an interruption, and regenerating all figures
and tables are listed in Appendix~\ref{app:repro}.

\FloatBarrier
\section{Experimental Setup}
\label{sec:setup}

Table~\ref{tab:protocol} lists the frozen protocol shared by both experiments
(mirrored in \texttt{data/experiment\_protocol.json}).
The same pipeline code is run on CIFAR-10 and CIFAR-100
($32\times32$)~\cite{krizhevsky2009learning} with dataset-specific GPU accuracy
thresholds (0.40 and 0.20, respectively, reflecting the harder 100-way task)
and separate artifact trees.
Both runs use a single Samsung Galaxy Tab SM-P613 (Android~14) connected over
USB debugging---no emulator---the base model
\texttt{ABrain/NNGPT-UniqueArch-Rag}~\cite{nngpt_uniquearch_rag}, seed~42 for
generation and data shuffling, and 20 generated models per cycle.
QLoRA fine-tuning and GPU evaluation run on a Linux workstation with an NVIDIA
GPU.

Per cycle we report best and average first-epoch GPU accuracy, the best mobile
score, mean quantized accuracy, and the number of \emph{valid} mobile models
(successful TFLite export with positive measured latency); 95\% confidence
intervals use Student's $t$ across the models within a cycle
($n \approx 15$--20).
All aggregated metrics bundled with this paper
(\texttt{data/progress\_metrics\_cifar10.json},
\texttt{data/progress\_metrics\_cifar100.json}) are produced directly from the
per-cycle records by the plotting script described in
Appendix~\ref{app:repro}.

\begin{table}[tbp]
\centering
\caption{Frozen experimental protocol (shared across both benchmarks). Dataset-specific thresholds and artifact roots differ as noted.}
\label{tab:protocol}
\small
\begin{tabular}{@{}p{0.34\textwidth}p{0.60\textwidth}@{}}
\toprule
Parameter & Value \\
\midrule
Datasets & CIFAR-10 and CIFAR-100 ($32\times32$)~\cite{krizhevsky2009learning} \\
Mobile device & Samsung Galaxy Tab SM-P613 (Android~14) \\
Base LLM & \texttt{ABrain/NNGPT-UniqueArch-Rag}~\cite{kochnev2025nngpt,nngpt_uniquearch_rag} \\
LLM adaptation & QLoRA~\cite{dettmers2023qlora}, 3 epochs per cycle \\
Models per cycle & 20 \\
GPU training & 1 epoch per candidate \\
GPU accuracy threshold & 0.40 (CIFAR-10); 0.20 (CIFAR-100) \\
Mobile export & INT8 TensorFlow Lite~\cite{jacob2018quantization,tensorflowlite} \\
Mobile metric & quantized accuracy / best latency \\
Delegates & GPU, CPU, NNAPI (20 timed runs each) \\
Gate tolerance & 0.99 \\
Random seed & 42 \\
Cycles reported & 0--6 (CIFAR-10); 0--6 (CIFAR-100); 8 planned \\
Artifact roots & \texttt{out-cfar10-backup/}; \texttt{out/curation\_output/} \\
\bottomrule
\end{tabular}
\end{table}

\begin{table}[tbp]
\centering
\caption{Cross-dataset comparison (seed~42, same device and pipeline). ``Peak mobile'' is the best score over cycles~0--6; ``Peak cycle'' is the cycle index that achieved it.}
\label{tab:cross-dataset}
\small
\begin{tabular}{@{}lcccc@{}}
\toprule
Dataset & Baseline mobile (cycle~0) & Peak mobile & Peak cycle & Peak / baseline \\
\midrule
CIFAR-10 & $1.090\times10^{-7}$ & $2.786\times10^{-6}$ & 1 & 25.6$\times$ \\
CIFAR-100 & $9.657\times10^{-8}$ & $9.657\times10^{-8}$ & 0 & 1.0$\times$ \\
\bottomrule
\end{tabular}
\end{table}

\FloatBarrier
\section{Results}
\label{sec:results}

Table~\ref{tab:cross-dataset} contrasts the two benchmarks under the identical
pipeline, device, and gate; Sections~\ref{sec:results-cifar10}
and~\ref{sec:results-cifar100} report the per-cycle evidence.

\subsection{CIFAR-10}
\label{sec:results-cifar10}

\begin{figure}[tbp]
  \centering
  \begin{subfigure}[t]{0.48\textwidth}
    \centering
    \includegraphics[width=\linewidth]{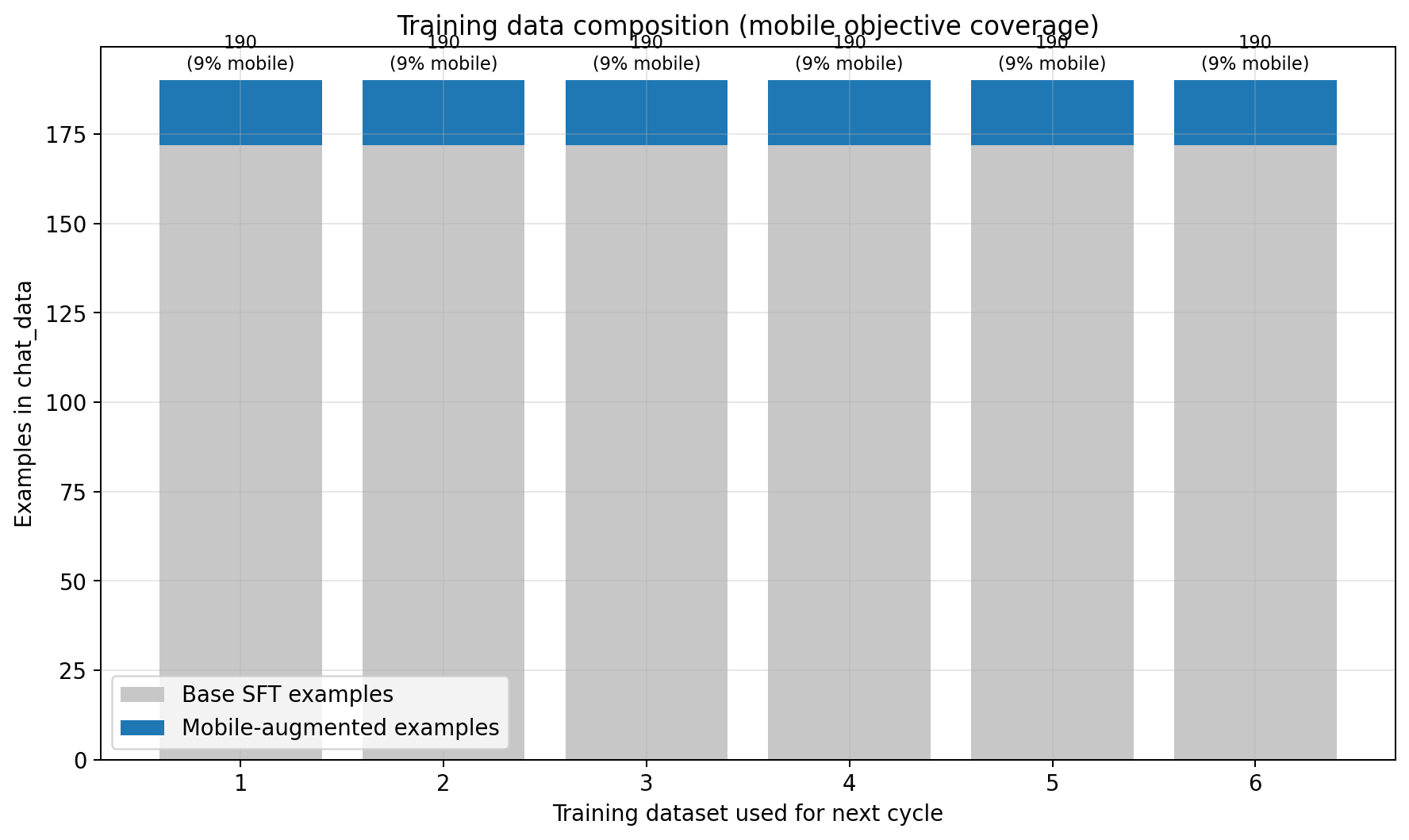}
    \caption{Training-pool size per cycle.}
    \label{fig:training-pool-cifar10}
  \end{subfigure}\hfill
  \begin{subfigure}[t]{0.48\textwidth}
    \centering
    \includegraphics[width=\linewidth]{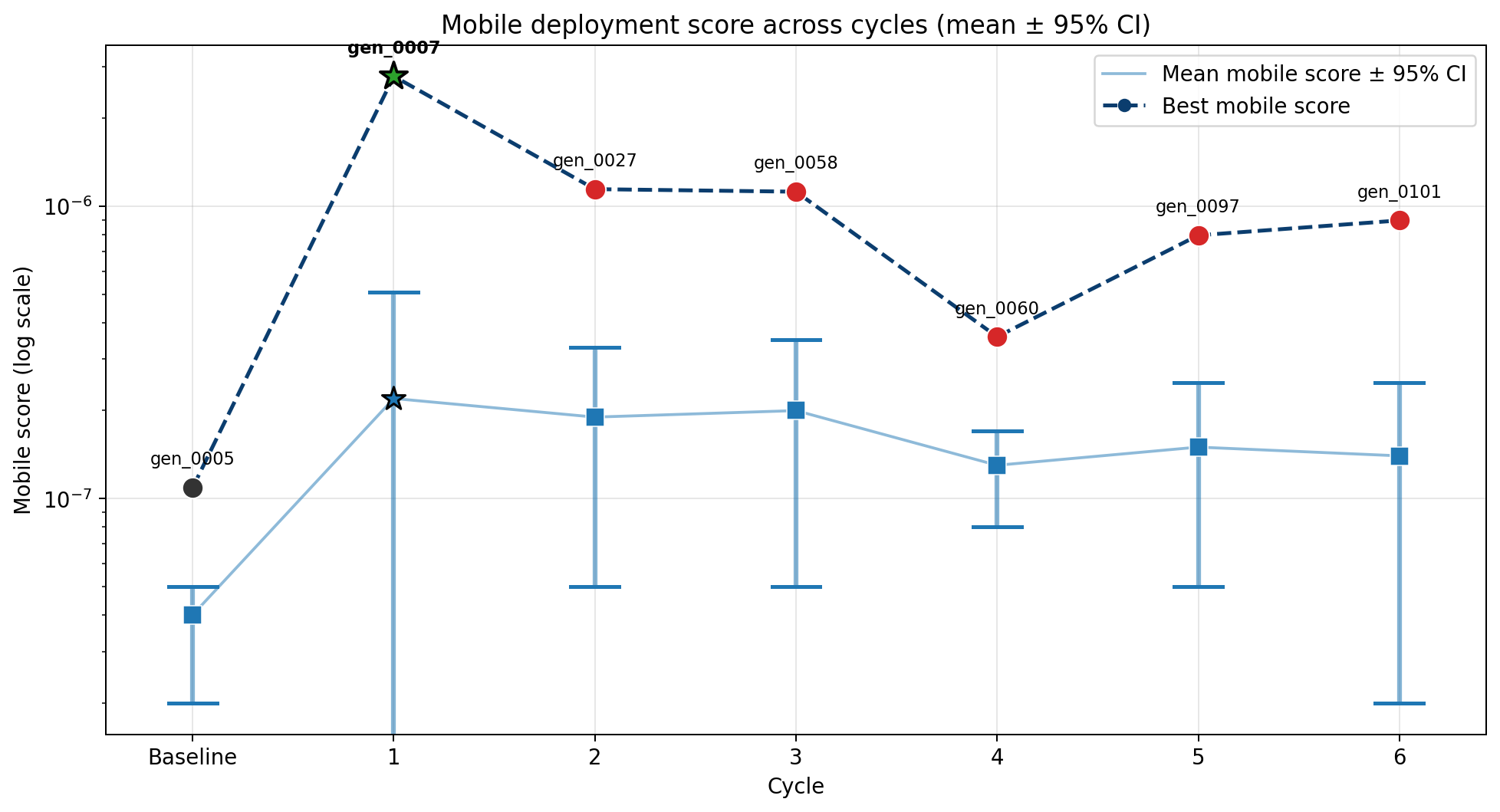}
    \caption{Best mobile score (log scale, 95\% CI).}
    \label{fig:mobile-score-cifar10}
  \end{subfigure}\\[4pt]
  \begin{subfigure}[t]{0.48\textwidth}
    \centering
    \includegraphics[width=\linewidth]{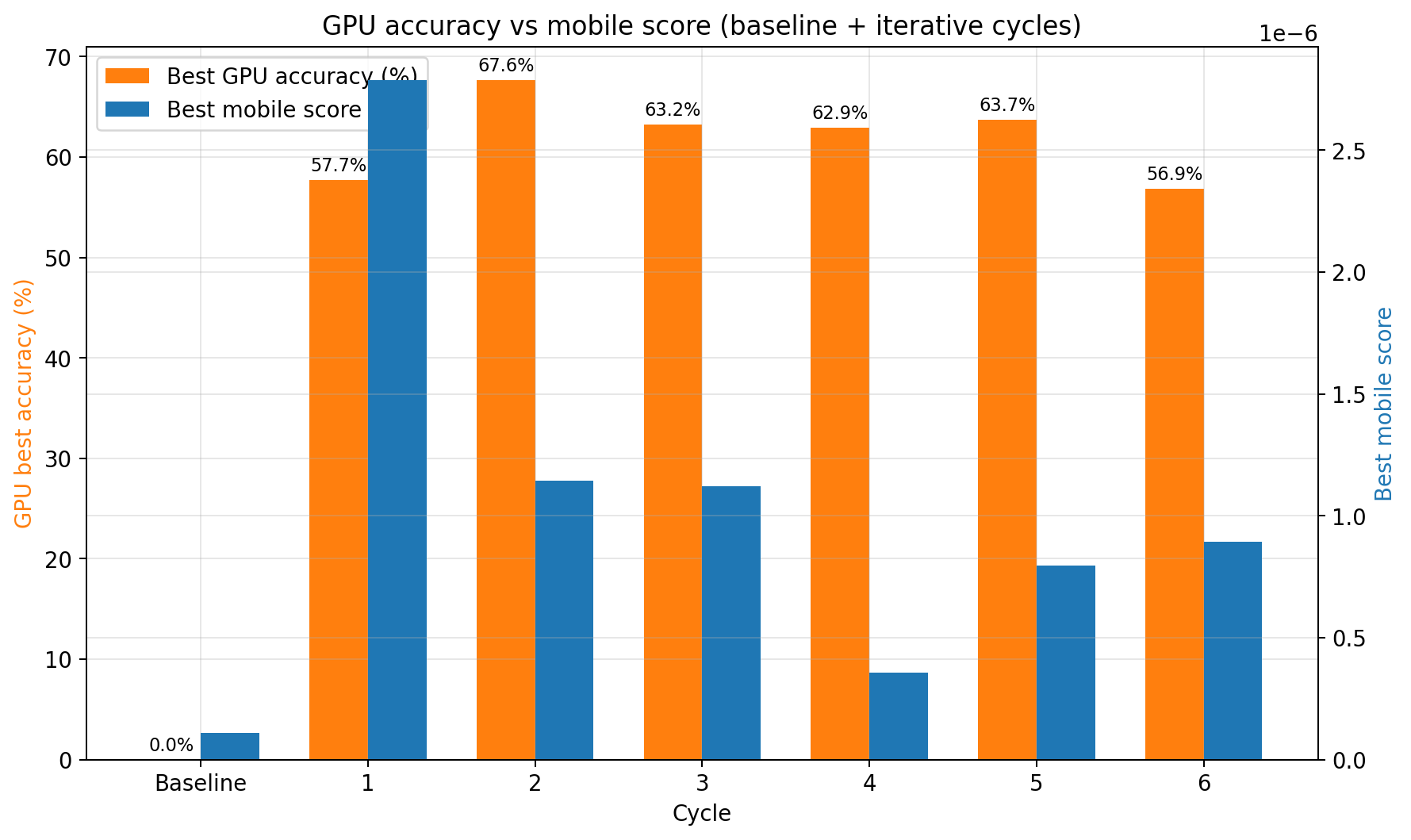}
    \caption{GPU-best vs.\ mobile-best score.}
    \label{fig:gpu-mobile-cifar10}
  \end{subfigure}\hfill
  \begin{subfigure}[t]{0.48\textwidth}
    \centering
    \includegraphics[width=\linewidth]{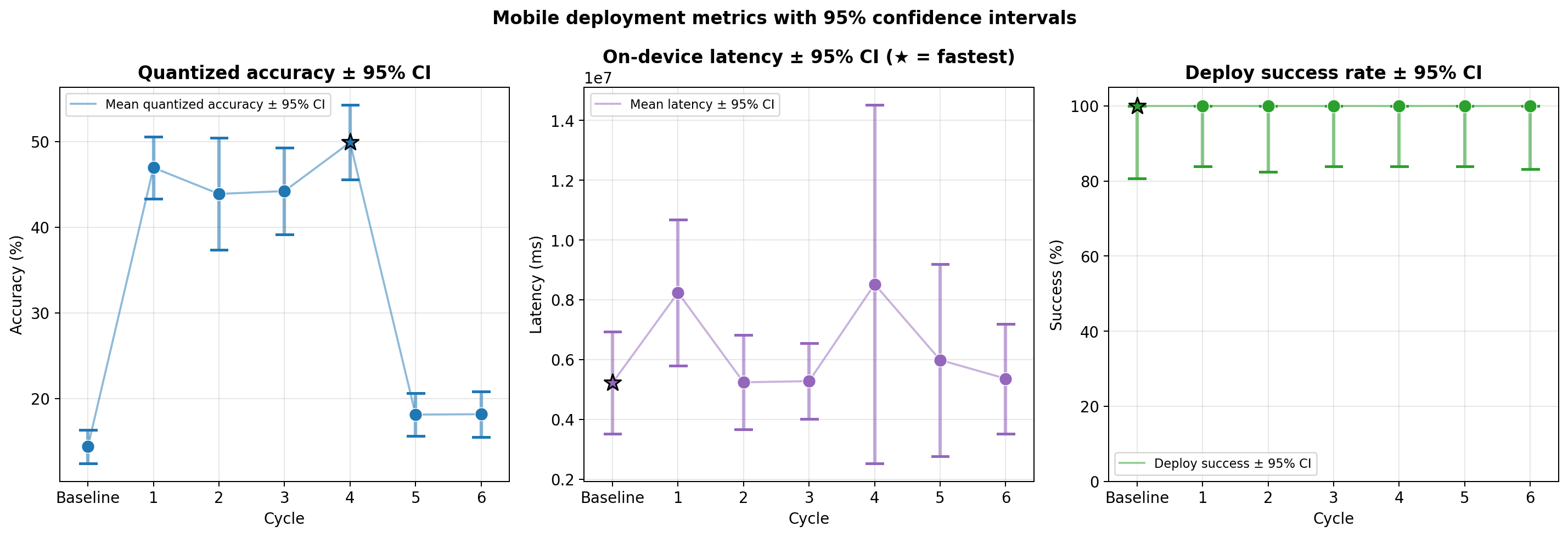}
    \caption{Mobile metrics with 95\% CIs.}
    \label{fig:mobile-ci-cifar10}
  \end{subfigure}
  \caption{CIFAR-10 per-cycle behavior. (a)~The pool grows only at the accepted
  cycle~1 (190 examples) and then remains flat. (b)~The cycle~1 mobile-score
  peak is never surpassed. (c)~GPU and mobile objectives diverge in later
  cycles. (d)~Quantized accuracy, latency, and deploy success with confidence
  intervals.}
  \label{fig:cifar10-panels}
\end{figure}

Table~\ref{tab:pipeline-summary-cifar10} and Figure~\ref{fig:cifar10-panels} report cycles~0--6.
The ungated baseline (cycle~0) achieves a best mobile score of
$1.09\times10^{-7}$ (model \texttt{gen\_0005}) with a mean quantized accuracy of
14.4\% across 16 valid models; aggregated GPU accuracy fields for this cycle are
zero in the archived summary because GPU logging was not backfilled into the
plot aggregate, whereas the mobile measurements are complete.
The first fine-tuned cycle transforms both objectives: GPU best accuracy rises
to 57.7\%, mean quantized accuracy reaches 46.9\%, and the best mobile score
improves to $2.79\times10^{-6}$---approximately $25.6\times$ the baseline.
The gate accepts this cycle, and 18 novel models expand the training pool to
190 examples (Figure~\ref{fig:cifar10-panels}a).

Every subsequent cycle fails the gate relative to the cycle~1 champion, even
though GPU best accuracy peaks at 67.6\% in cycle~2 and remains above 56\%
throughout.
Mean quantized accuracy stays near 44--50\% through cycle~4 but drops to
approximately 18\% in cycles~5--6 while GPU metrics remain strong, reproducing
at larger scale the divergence between workstation and on-device objectives
that motivates the gate.
Deployment robustness is unaffected: the deploy success rate is 100\% for all
cycles with valid samples (95\% CI lower bound $\geq 0.82$).

\begin{table}[tbp]
\centering
\caption{Mobile iterative fine-tuning pipeline summary (CIFAR-10, SM-P613, seed~42). \textbf{Bold green} cells mark the best value in each column (same rule as the thesis plots).}
\label{tab:pipeline-summary-cifar10}
\footnotesize
\setlength{\tabcolsep}{3.5pt}
\adjustbox{max width=\textwidth}{%
\begin{tabular}{@{}lrrllllr@{}}
\toprule
Cycle & GPU best & Mobile best & \multicolumn{1}{c}{Mean score (95\% CI)} & \multicolumn{1}{c}{Mean quant acc (95\% CI)} & Gate & Del. & Valid \\
\midrule
Baseline & 0.0\% & $1.090\times10^{-7}$ & $4.00\times10^{-8} \pm 1.00\times10^{-8}$ & 14.4\% $\pm$ 2.0\% & baseline & cpu & 16 \\
1 & 57.7\% & \cellcolor{tablebest}\textbf{$2.786\times10^{-6}$} & \cellcolor{tablebest}\textbf{$2.20\times10^{-7} \pm 2.90\times10^{-7}$} & 46.9\% $\pm$ 3.6\% & \cellcolor{tablebest}\textbf{accepted} & gpu & \cellcolor{tablebest}\textbf{20} \\
2 & \cellcolor{tablebest}\textbf{67.6\%} & $1.145\times10^{-6}$ & $1.90\times10^{-7} \pm 1.40\times10^{-7}$ & 43.9\% $\pm$ 6.5\% & rejected & cpu & 18 \\
3 & 63.2\% & $1.122\times10^{-6}$ & $2.00\times10^{-7} \pm 1.50\times10^{-7}$ & 44.2\% $\pm$ 5.0\% & rejected & gpu & \cellcolor{tablebest}\textbf{20} \\
4 & 62.9\% & $3.578\times10^{-7}$ & $1.30\times10^{-7} \pm 5.00\times10^{-8}$ & \cellcolor{tablebest}\textbf{49.9\% $\pm$ 4.4\%} & rejected & cpu & \cellcolor{tablebest}\textbf{20} \\
5 & 63.7\% & $7.957\times10^{-7}$ & $1.50\times10^{-7} \pm 1.00\times10^{-7}$ & 18.1\% $\pm$ 2.5\% & rejected & cpu & \cellcolor{tablebest}\textbf{20} \\
6 & 56.9\% & $8.946\times10^{-7}$ & $1.40\times10^{-7} \pm 1.10\times10^{-7}$ & 18.2\% $\pm$ 2.7\% & rejected & cpu & 19 \\
\bottomrule
\end{tabular}
}
\end{table}

\subsection{CIFAR-100}
\label{sec:results-cifar100}

\begin{figure}[tbp]
  \centering
  \begin{subfigure}[t]{0.48\textwidth}
    \centering
    \includegraphics[width=\linewidth]{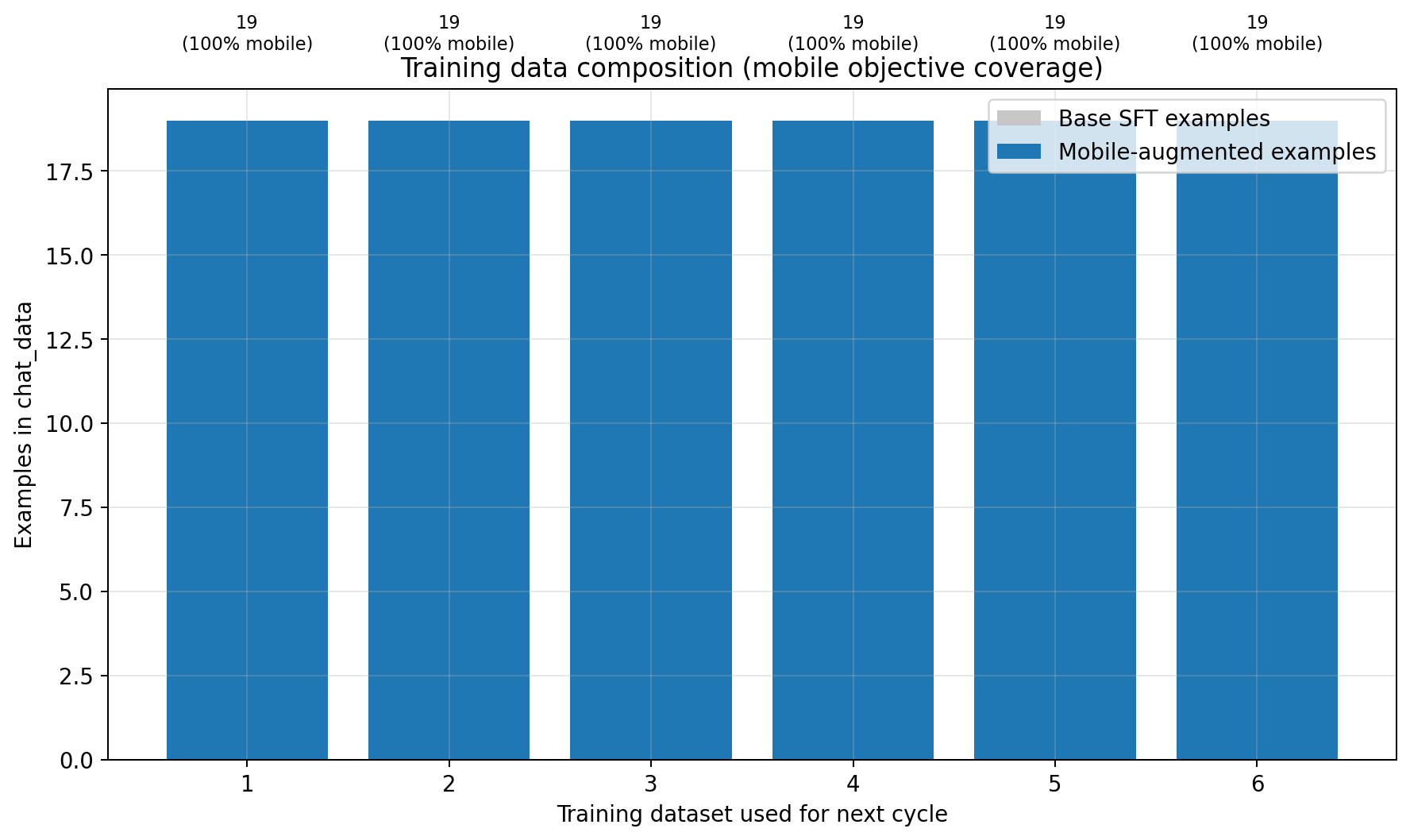}
    \caption{Training-pool size per cycle.}
    \label{fig:training-pool-cifar100}
  \end{subfigure}\hfill
  \begin{subfigure}[t]{0.48\textwidth}
    \centering
    \includegraphics[width=\linewidth]{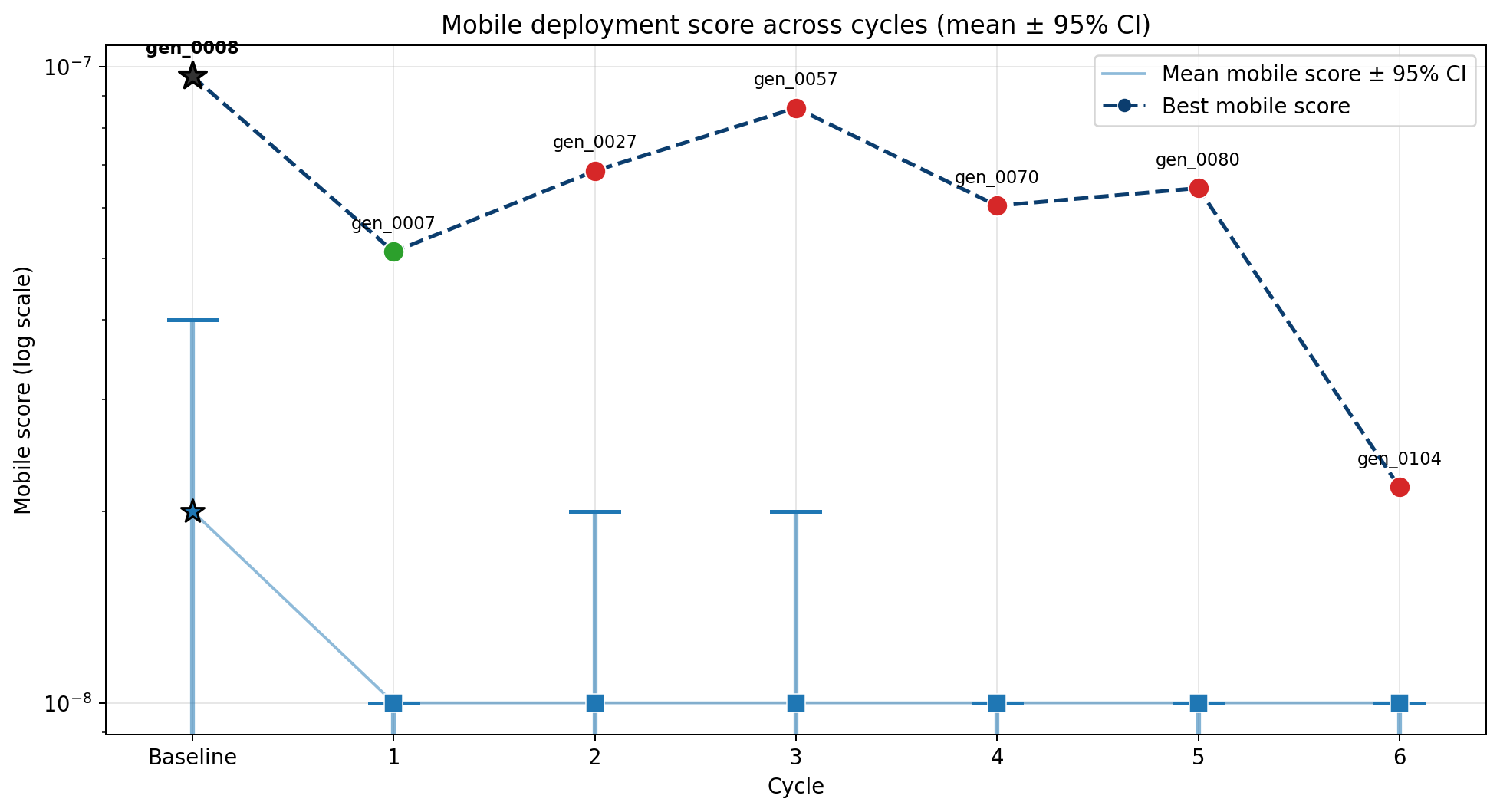}
    \caption{Best mobile score (log scale, 95\% CI).}
    \label{fig:mobile-score-cifar100}
  \end{subfigure}\\[4pt]
  \begin{subfigure}[t]{0.48\textwidth}
    \centering
    \includegraphics[width=\linewidth]{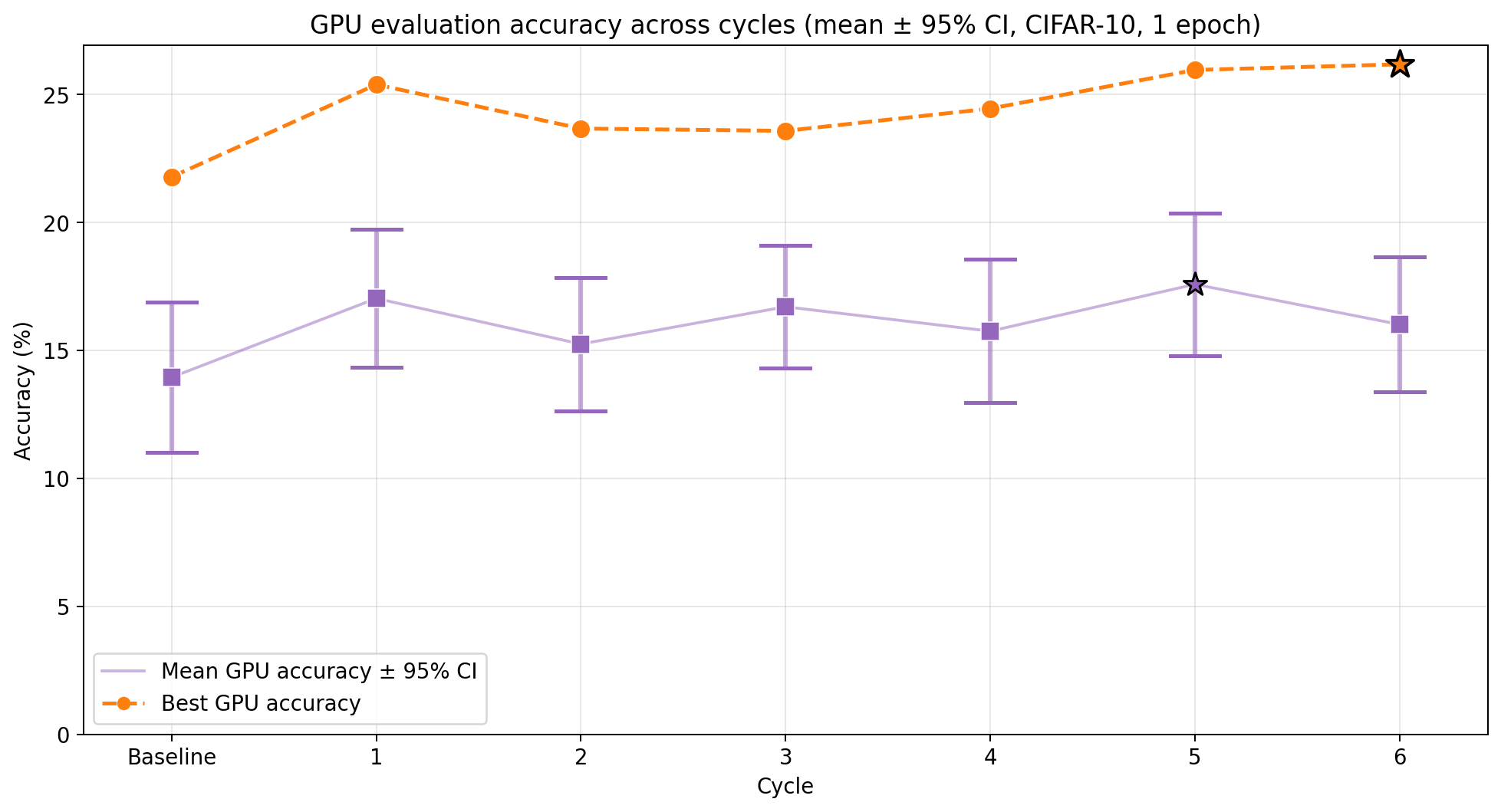}
    \caption{GPU first-epoch accuracy (95\% CI).}
    \label{fig:gpu-accuracy-cifar100}
  \end{subfigure}\hfill
  \begin{subfigure}[t]{0.48\textwidth}
    \centering
    \includegraphics[width=\linewidth]{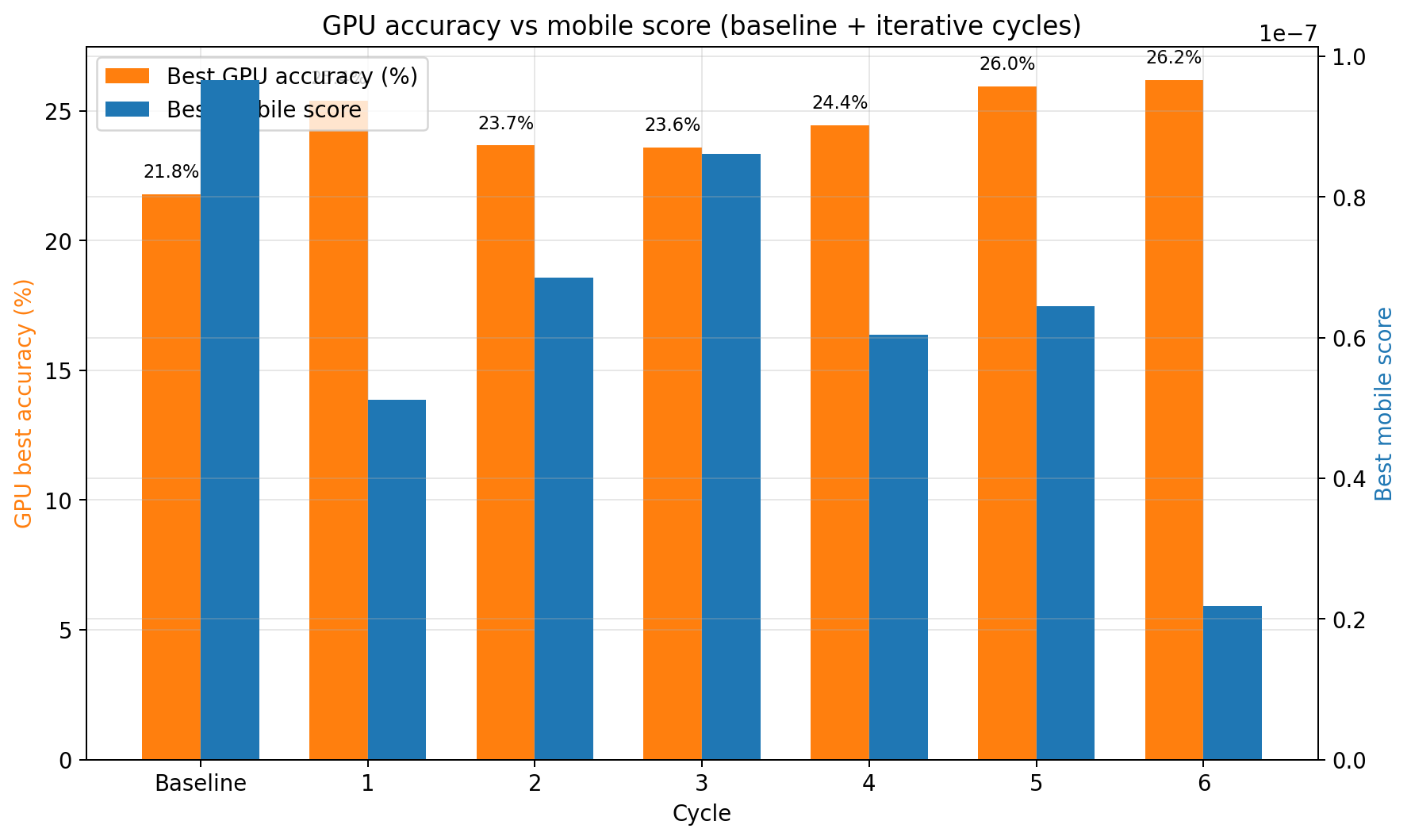}
    \caption{GPU-best vs.\ mobile-best score.}
    \label{fig:gpu-mobile-cifar100}
  \end{subfigure}
  \caption{CIFAR-100 per-cycle behavior. (a)~The pool grows only at cycle~1
  (19 examples) and then remains frozen. (b)~The cycle~0 baseline remains the
  deployment peak. (c)~GPU accuracy improves across cycles while (d)~the mobile
  objective does not follow.}
  \label{fig:cifar100-panels}
\end{figure}

\begin{figure}[tbp]
  \centering
  \adjustbox{max width=0.72\textwidth,keepaspectratio}{%
    \includegraphics{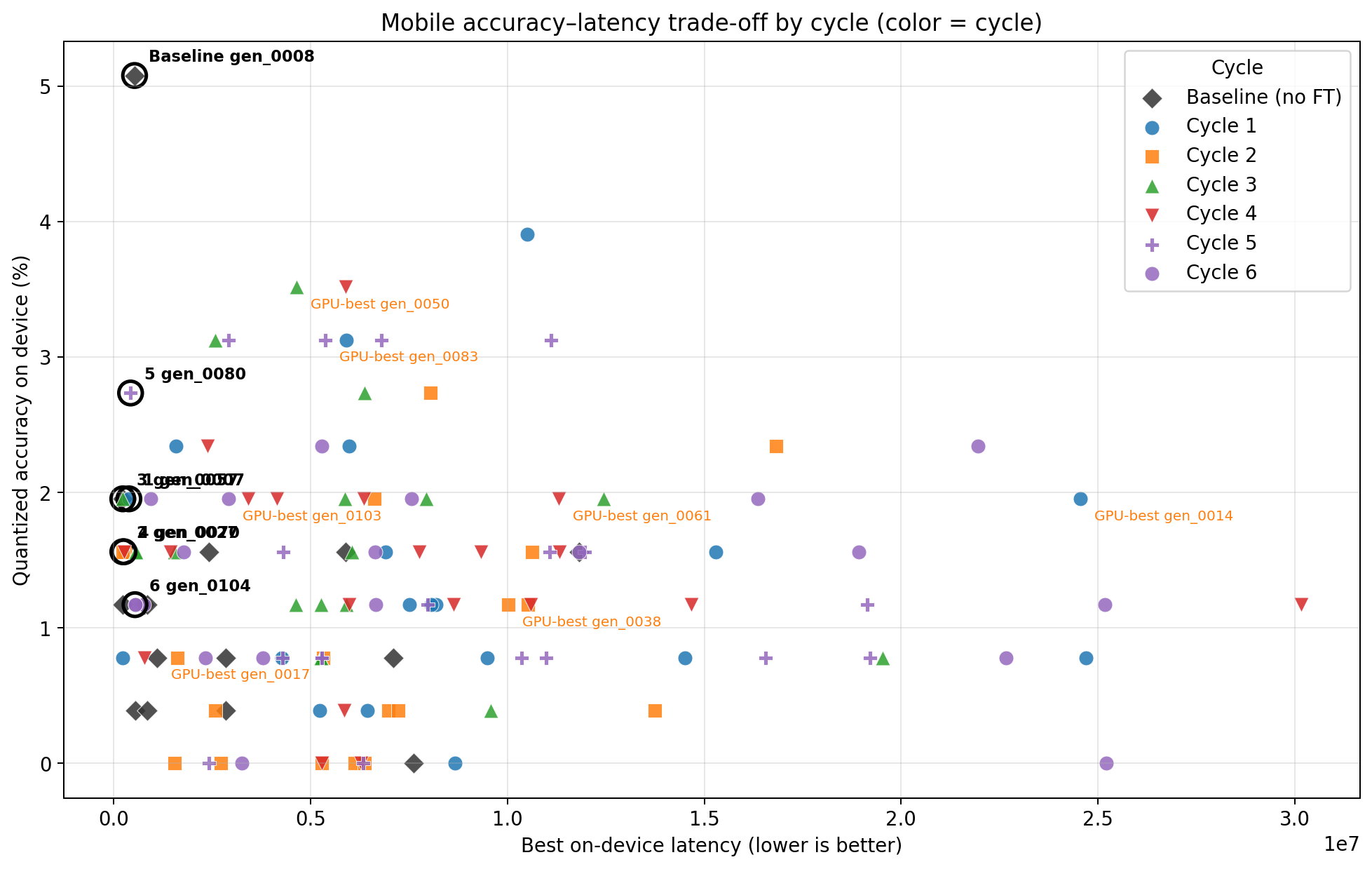}%
  }
  \caption{CIFAR-100: on-device quantized accuracy versus latency, colored by
  cycle. No cycle produces a candidate that dominates the cycle~0 champion in
  both dimensions.}
  \label{fig:mobile-scatter-cifar100}
\end{figure}

Table~\ref{tab:pipeline-summary-cifar100} and Figures~\ref{fig:cifar100-panels} and~\ref{fig:mobile-scatter-cifar100} report cycles~0--6.
On this harder 100-way task the pre-fine-tuning baseline is never beaten on the
deployment objective: cycle~0 attains a GPU best accuracy of 21.78\% and the
overall best mobile score of $9.66\times10^{-8}$ (model \texttt{gen\_0008}),
with a mean quantized accuracy of 1.25\% over 15 valid models.
Cycle~1 raises GPU best accuracy to 25.39\% and is accepted by the gate---as the
first iterative round it has no prior gated reference---adding 19 novel chat
examples, yet its best mobile score ($5.12\times10^{-8}$) already falls short of
the baseline.

All later cycles are compared against $0.99 \times 9.66\times10^{-8}$ and are
rejected; cycle~3 comes closest at $8.61\times10^{-8}$.
GPU best accuracy nevertheless climbs to 26.18\% by cycle~6 while mean quantized
accuracy stays in the 1.3--1.6\% range, and the training pool consequently
remains frozen at 19 examples after cycle~1
(Figure~\ref{fig:cifar100-panels}a).
As on CIFAR-10, deployability is robust: the deploy success rate is at or near
100\% for all cycles with valid samples, confirming that the pipeline reliably
produces exportable, runnable graphs even when their quantized accuracy is low.

\begin{table}[tbp]
\centering
\caption{Mobile iterative fine-tuning pipeline summary (CIFAR-100, SM-P613, seed~42). \textbf{Bold green} cells mark the best value in each column (same rule as the thesis plots).}
\label{tab:pipeline-summary-cifar100}
\footnotesize
\setlength{\tabcolsep}{3.5pt}
\adjustbox{max width=\textwidth}{%
\begin{tabular}{@{}lrrllllr@{}}
\toprule
Cycle & GPU best & Mobile best & \multicolumn{1}{c}{Mean score (95\% CI)} & \multicolumn{1}{c}{Mean quant acc (95\% CI)} & Gate & Del. & Valid \\
\midrule
Baseline & 21.8\% & \cellcolor{tablebest}\textbf{$9.657\times10^{-8}$} & \cellcolor{tablebest}\textbf{$2.00\times10^{-8} \pm 2.00\times10^{-8}$} & 1.2\% $\pm$ 0.7\% & baseline & cpu & 15 \\
1 & 25.4\% & $5.115\times10^{-8}$ & $1.00\times10^{-8} \pm 1.00\times10^{-8}$ & 1.4\% $\pm$ 0.5\% & \cellcolor{tablebest}\textbf{accepted} & cpu & 19 \\
2 & 23.7\% & $6.856\times10^{-8}$ & $1.00\times10^{-8} \pm 1.00\times10^{-8}$ & 0.9\% $\pm$ 0.4\% & rejected & gpu & 19 \\
3 & 23.6\% & $8.613\times10^{-8}$ & $1.00\times10^{-8} \pm 1.00\times10^{-8}$ & \cellcolor{tablebest}\textbf{1.6\% $\pm$ 0.4\%} & rejected & cpu & 19 \\
4 & 24.4\% & $6.044\times10^{-8}$ & $1.00\times10^{-8} \pm 1.00\times10^{-8}$ & 1.4\% $\pm$ 0.4\% & rejected & gpu & \cellcolor{tablebest}\textbf{20} \\
5 & 26.0\% & $6.442\times10^{-8}$ & $1.00\times10^{-8} \pm 1.00\times10^{-8}$ & 1.5\% $\pm$ 0.5\% & rejected & gpu & 19 \\
6 & \cellcolor{tablebest}\textbf{26.2\%} & $2.183\times10^{-8}$ & $1.00\times10^{-8} \pm 0.00\times10^{0}$ & 1.3\% $\pm$ 0.3\% & rejected & cpu & \cellcolor{tablebest}\textbf{20} \\
\bottomrule
\end{tabular}
}
\end{table}

\subsection{Cross-dataset comparison}
\label{sec:results-compare}

The two experiments bracket the behavior of closed-loop mobile feedback.
On the easier ten-way task, a single gate-accepted round of fine-tuning yields a
large deployment gain ($25.6\times$ over baseline) and a rich training pool of
190 examples; on the 100-way task, the same procedure never surpasses the
pre-fine-tuning baseline on-device, and augmentation stalls at 19 examples.
In both runs GPU best accuracy continues to improve in cycles the gate rejects,
which demonstrates that workstation metrics alone would give a misleading
picture of progress: without on-device measurement, both campaigns would appear
to be succeeding.
The consistent 100\% deploy success rate across cycles further indicates that
the generative model reliably emits exportable graphs; the binding constraint is
quantized accuracy per unit latency, not deployability itself.

\FloatBarrier
\section{Discussion}
\label{sec:discussion}

The two benchmarks tell complementary stories about closed-loop mobile
feedback.
On CIFAR-10, one accepted round of QLoRA is highly effective for the deployment
objective---the mobile score improves by roughly $25.6\times$, quantized
accuracy approaches 47\%, and the training pool grows to 190 examples---whereas
on CIFAR-100 the identical code and device leave the pre-fine-tuning baseline
unbeaten and the pool frozen at 19 examples.
Task difficulty thus changes not only absolute accuracy but whether iterative
generative fine-tuning helps or hurts on-device quality, an effect that would
be invisible in a single-dataset study.

The stalling behavior after the first accepted round has a clear mechanism.
GPU fine-tuning optimizes first-epoch workstation accuracy, a proxy that does
not track INT8 on-device performance: in the late cycles of both runs, GPU best
accuracy remains high (above 56\% on CIFAR-10, rising to 26.2\% on CIFAR-100)
while mean quantized accuracy collapses or stagnates.
The non-decreasing gate is doing its job---it prevents regressions on the
deployment objective---but once it rejects a cycle, augmentation stops and the
language model keeps training on a fixed pool, so the loop can no longer
recover through data.
A stricter variant that requires beating the cycle~0 reference before any
augmentation would have kept the CIFAR-100 seed corpus unchanged; softer
alternatives include multi-objective gates with separate accuracy and latency
floors, or admitting a bounded number of examples from rejected cycles to keep
the pool from freezing.

Several limitations qualify these findings.
Both studies use a single tablet (SM-P613), a single reported seed per dataset,
and an incomplete eight-cycle budget; the CIFAR-10 cycle~0 GPU aggregate is
zero in the bundled JSON although its mobile measurements are complete.
The mobile score of Eq.~\eqref{eq:mobile-score} deliberately collapses accuracy
and latency into one ratio, which favors fast low-accuracy models when absolute
accuracies are small, as on CIFAR-100.
Future work includes multi-seed aggregation, quantization-aware generation
prompts, and fine-tuning objectives that weight mobile metrics directly rather
than filtering after the fact.

\section{Conclusion}
\label{sec:conclusion}

We presented an automated pipeline that closes the loop from QLoRA fine-tuning
of an architecture-generating LLM through GPU evaluation, INT8 TensorFlow Lite
export, and physical-device benchmarking to gated augmentation of the training
corpus, and evaluated it under a frozen protocol on both CIFAR-10 and
CIFAR-100.
The paired experiments show that closed-loop fine-tuning can deliver large
on-device gains when the task is tractable, that harder classification can
leave the pre-fine-tuning baseline unbeaten, and that in either regime rising
GPU accuracy is not evidence of deployment progress.
These results argue for mobile-native objectives inside generative AutoML
loops, and the released artifacts---per-cycle metrics with confidence
intervals, figures, and full reproduction commands
(Appendix~\ref{app:repro})---provide a baseline for such work.

\section*{Acknowledgments}
We thank the NNGPT and LEMUR teams at the University of W\"urzburg for models,
datasets, and benchmark infrastructure.

\bibliographystyle{plain}
\bibliography{references}

\appendix
\section{Reproduction Commands}
\label{app:repro}

All commands run from the repository root with dependencies installed, base LLM
weights under \texttt{out/llm/}, and the SM-P613 connected with USB debugging.
Replace \texttt{DATASET} with \texttt{cifar-10} or \texttt{cifar-100} and
\texttt{THRESH} with 0.40 or 0.20, respectively.
CIFAR-10 artifacts live under \texttt{out-cfar10-backup/curation\_output/};
CIFAR-100 artifacts under \texttt{out/curation\_output/}.

\paragraph{Baseline cycle 0 (pre-QLoRA reference).}
\begin{small}
\begin{verbatim}
python3 -m ab.gpt.TuneNNGen --run_iterative_pipeline \
  --mobile_deployment --baseline_only \
  --dataset DATASET --save_eval_checkpoint \
  --llm_conf nngpt_unique_arch_rag.json \
  --accuracy_threshold THRESH --seed 42
\end{verbatim}
\end{small}

\paragraph{Full iterative run (cycles 1--8).}
\begin{small}
\begin{verbatim}
python3 -m ab.gpt.TuneNNGen --run_iterative_pipeline \
  --mobile_deployment --dataset DATASET \
  --save_eval_checkpoint \
  --llm_conf nngpt_unique_arch_rag.json \
  --accuracy_threshold THRESH --cycles 8 \
  --models_per_cycle 20 --seed 42
\end{verbatim}
\end{small}

\paragraph{Resume after interruption.}
Existing per-cycle checkpoints and generation results are reused; completed
stages are not repeated.
\begin{small}
\begin{verbatim}
python3 -m ab.gpt.TuneNNGen --run_iterative_pipeline \
  --mobile_deployment --dataset DATASET \
  --save_eval_checkpoint \
  --llm_conf nngpt_unique_arch_rag.json \
  --accuracy_threshold THRESH --seed 42 \
  --resume_from_cycle N --cycles 8
\end{verbatim}
\end{small}

\paragraph{Regenerate figures and tables.}
\begin{small}
\begin{verbatim}
# CIFAR-10 (backup run)
python test/plot_finetune_mobile_cycles.py \
  --curation-root out-cfar10-backup/curation_output \
  --out-dir out-cfar10-backup/curation_output/thesis_progress
cp out-cfar10-backup/curation_output/thesis_progress/*.png \
   paper_mobile_deployment/figures/cifar10/
cp out-cfar10-backup/curation_output/thesis_progress/progress_metrics.json \
   paper_mobile_deployment/data/progress_metrics_cifar10.json

# CIFAR-100 (active out/)
python test/plot_finetune_mobile_cycles.py \
  --curation-root out/curation_output \
  --out-dir out/curation_output/thesis_progress
cp out/curation_output/thesis_progress/*.png \
   paper_mobile_deployment/figures/cifar100/
cp out/curation_output/thesis_progress/progress_metrics.json \
   paper_mobile_deployment/data/progress_metrics_cifar100.json

python paper_mobile_deployment/scripts/generate_tables.py
\end{verbatim}
\end{small}

\paragraph{Compile this document.}
\begin{small}
\begin{verbatim}
cd paper_mobile_deployment && make
\end{verbatim}
\end{small}
The folder is self-contained for Overleaf (main file \texttt{main.tex}).

\end{document}